\documentclass{article}

\usepackage{arxiv}

\usepackage[utf8]{inputenc} 
\usepackage[T1]{fontenc}    
\usepackage{hyperref}       
\usepackage{url}            
\usepackage{booktabs}       
\usepackage{amsfonts}       
\usepackage{nicefrac}       
\usepackage{microtype}      
\usepackage{lipsum}
\usepackage{graphicx}
\graphicspath{ {./images/} }
\usepackage{caption} 
\usepackage{natbib}
\setcitestyle{numbers,square}
\usepackage{url,hyperref,lineno,microtype,subcaption}
\usepackage[onehalfspacing]{setspace}
\usepackage{adjustbox}

\usepackage{makecell}

\title{Detection of fake news on CoViD-19 \\on Web Search Engines}

\author{V. Mazzeo\thanks{Department of Physics and Astronomy “E. Majorana”, University of Catania, Italy}\\ \texttt{valeria.mazzeo@phd.unict.it}\\ \texttt{valesdn@gmail.com} 
\and \textbf{A. Rapisarda}\thanks{Department of Physics and Astronomy “E. Majorana”, University of Catania and INFN Sezione di Catania, Italy;  Complexity Science Hub Vienna (CSH), Austria;}\\ \texttt{andrea.rapisarda@unict.it} 
\and \textbf{G. Giuffrida}\thanks{Department of Political and Social Sciences (DSPS), University of Catania, Italy}\\ \texttt{giovanni.giuffrida@unict.it}}

\begin{document}
\onecolumn


\maketitle
\begin{abstract}

In early January 2020, after China reported the first cases of the new coronavirus (SARS-CoV-2) in the city of Wuhan, unreliable and not fully accurate information has started spreading faster than the virus itself. Alongside this pandemic, people have experienced a parallel \textit{infodemic}, i.e., an overabundance of information, some of which misleading or even harmful, that has widely spread around the globe. Although Social Media are increasingly being used as information source, Web Search Engines, like Google or Yahoo!, still represent a powerful and trustworthy resource for finding information on the Web. This is due to their capability to capture the largest amount of information,  helping users quickly identify the most relevant, useful, although not always the most reliable, results for their search queries.\\ This study aims to detect potential misleading and fake contents by capturing and analysing textual information, which flow through Search Engines. By using a real-world dataset associated with recent CoViD-19 pandemic, we first apply re-sampling techniques for class imbalance, then we use existing Machine Learning algorithms for classification of not reliable news. By extracting lexical and host-based features of associated Uniform Resource Locators (URLs) for news articles, we show that the proposed methods, so common in phishing and malicious URLs detection, can improve the efficiency and performance of classifiers. \\ Based on these findings, we suggest that the use of both textual and URLs features can improve the effectiveness of fake news detection methods.

\footnotesize
\textbf{Keywords:} fake news, covid-19, search engine, machine learning, class imbalance, phishing 
\end{abstract}

\section{Introduction}
\label{0}
The reliability and credibility of both information source and information itself have emerged as a global issue in contemporary society \citep{Greer, Xianjin}. Undoubtedly, in the last decades Social Media have revolutionised the way in which information spreads across the Web and, more generally, the world \citep{Chou, Breland}, by allowing users to freely share content faster than traditional news sources. The fact that content spreads so quickly and easily across platforms suggests that people (and algorithms behind the platforms) are potentially vulnerable to misinformation, hoaxes, biases and low-credibility contents which are daily shared, accidentally or intentionally. \\
The problem of spreading misinformation, however, affects not only the Social Media platforms, but also the World Wide Web. In fact, every time people enter a search query on Web Search Engines (WSE), like \href{https://www.google.com/?client=safari}{Google} or \href{https://www.bing.com}{Bing}, they can view and potentially access hundreds, or thousands, of web pages with helpful information, sometimes potentially misleading. 
Meta title tags displayed on \href{https://developers.google.com/search/docs/advanced/appearance/good-titles-snippets}{Search Engine Results Pages} (SERPs) \citep{Page} represent then a crucial factor in helping user understand pages' content, being the user's first experience of a website. \\Although people tend more likely to view and click on the first results on the first page because of Search Engines' rank algorithms, e.g., PageRank (Page, 1998), which show the most relevant information in response to a specific query, a good title can be the \textit{make-or-break} factor which brings users to click on that link and read on \citep{Manjesh,Bourgonje}. \\
Despite the systematic and significant response efforts and fact-checking against misinformation mobilised by both \href{https://www.forbes.com/sites/bernardmarr/2020/03/27/finding-the-truth-about-covid-19-how-facebook-twitter-and-instagram-are-tackling-fake-news/?sh=7d38e7e51977}{Social Media and Media companies}, fake news still persist due to the vast volume of online content, which leads people to see and share information that is partly, or completely, misleading.
Previous and recent studies have almost exclusively focused on data from Social Media (e.g., \href{https://twitter.com}{Twitter})\citep{Aldwairi}, fact-checking or reliable websites (e.g., \href{https://www.snopes.com}{Snopes.com}, \href{https://www.politifact.com}{PolitiFact.com})\citep{Elhadad}, or existing datasets\citep{Vasu} which have the benefit to be cost-efficient.\\
Due to the current difficult and unprecedented situation with the CoViD-19 pandemic, never seen in the modern era \cite{Reuters}, people have asked many questions about the novel coronavirus, such as the origin of the disease,  treatment, prevention, cure, and transmission from or to pets, to face these challenges, while staying informed and safe.
In this study, we focus on news displayed by Web Search Engines, since they represent the best tools for bringing up answers to people’s current questions, extracting information related to CoViD-19 outbreak, and proposing an approach based on both textual and Uniform Resource Locator (URL) features to analyse and detect whether news are fake/misleading or reliable (real). 
The contribution of our work can be summarised as follows: 
\begin{itemize}
	\item we use real-world data from WSE, analysing both textual data (meta titles and descriptions) and URLs information, by extracting features representations; 
	\item since most of the previous works on fake news detection were focused on classifier enhancements, not engaging in feature engineering, in this document we want to provide a new direction for the classification of fake news, proposing an integration of the most commonly used features in fake news detection and features that play an important role in the malicious URL detection. The purpose of feature engineering is indeed  to feed the original data and provide new and meaningful feature representations to improve machine learning (ML) algorithms for classification. Currently, the problem of detecting fake news via URL has not been well and sufficiently addressed. Several studies focused on fake news detection via ML in Social Networks \citep{Madani} have looked at the presence of URLs in the user’s published content \citep{Helmstetter}, without generally performing further analysis on the source of information or extracting other potential relevant URL-based features (features that are, indeed, more common in malicious URLs/phishing detection classifiers). Although, in the past, the usage of URLs in a post/news could have represented a useful parameter for enhancing and improving ML classifiers' performance, nowadays this could result not enough for differentiating a good source from a bad one in terms of information credibility without a URL-based feature engineering approach. In fact, the more ML techniques have evolved over time, the more schemes for spreading fake news have changed;
	\item we apply re-sampling techniques, such as under-sampling and over-sampling, due to the class imbalance of the real-world dataset \citep{Desuky, Al2}. Disproportion between classes still represents an open issue and challenge for researchers focused on classification problems. In a typical news dataset, the number of fake news are likely to be very few compared to the number of real ones, and this fact makes the positive class (fake news) very small compared to the negative class (real news). This imbalance between the two classes would likely make classifiers biased towards the majority class leading to classify all the instances in the dataset as belonging to the majority class; 
	\item we compare different ML algorithms (Support Vector Machine, Stochastic Gradient Descent, Logistic Regression, Na\"ive Bayes, and Random Forest), based on their performance.
Since we deal with imbalanced data, we evaluate the models looking at $F_1$ score and Recall metrics, and not at predictive accuracy, as the latter represents a misleading indicator which reflects the underlying class distributions \citep{Sokolova, Lee}. 
\end{itemize}
The paper is structured as follows: Section \ref{2} introduces our material and methodology; Section \ref{3} describes the results of our experimentation along with their evaluation. In Section \ref{4} we summarise our key findings and give an interpretation of them, also by discussing the implications. Finally, Section \ref{5} draws the conclusions, giving some prospective points for future work.

\section{Material and Methods}
\label{2}
\subsection{Data collection}
\label{2a}
\begin{figure}[!htpb]
        \subfloat(A){\includegraphics[width=0.9\textwidth]{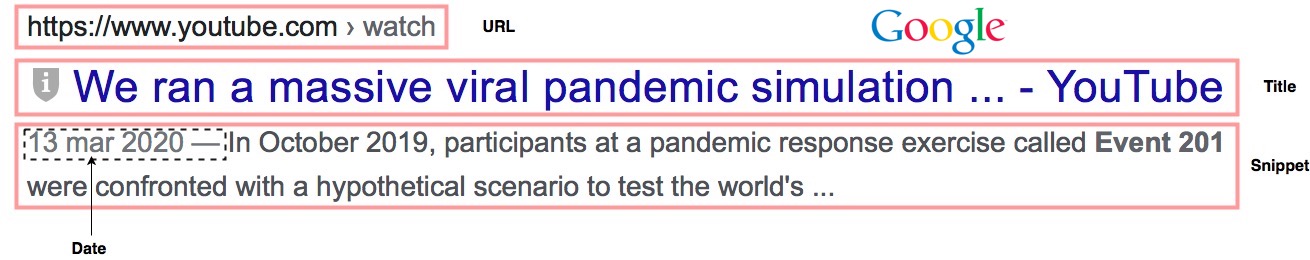}}
	    \subfloat(B){\includegraphics[width=0.42\textwidth]{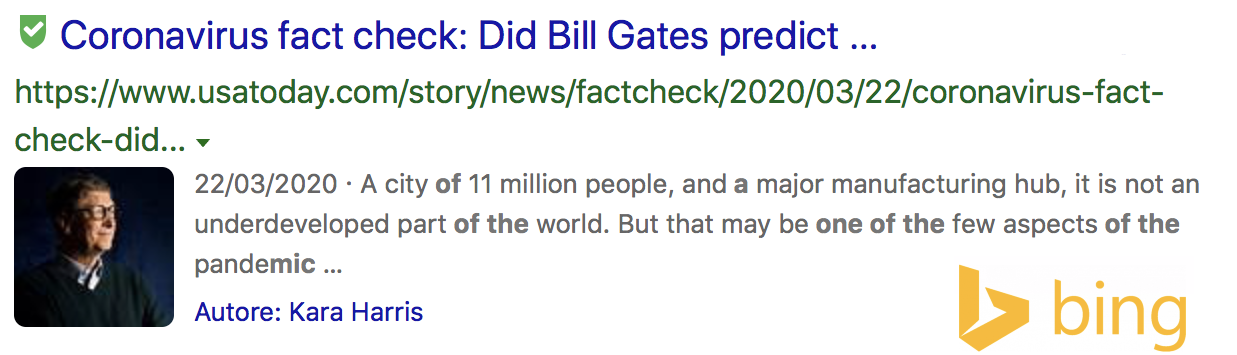}}
	    \subfloat(C){\includegraphics[width=0.48\textwidth]{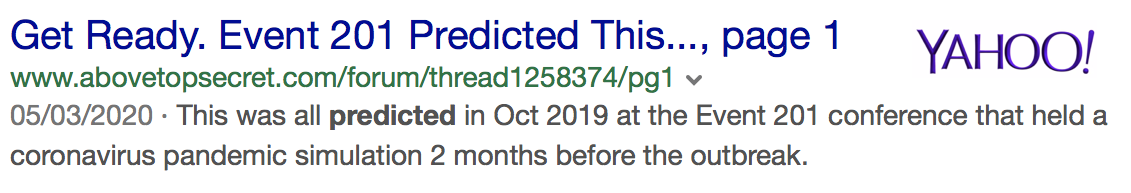}}
\caption{\small Example of search results on some of the most popular Web Search Engines: Google, Bing, and Yahoo!. The most important meta tags displayed by the Search Engine Results Pages, in response to user queries, are title, description (or snippet), date and cached URL.The sub-figure (A) is scaled up in order to highlight the meta tags (in red rectangles) extracted during data collection.}
\label{fig:1}
\end{figure}

We submitted several CoViD-19-related queries (\textbf{Table} \ref{tab:t1}) on a WSE. For each search result, we extracted metadata information, i.e., URL, title, meta description (or snippet), and date  (\textbf{Figure} \ref{fig:1}). 

\begin{table}[!htbp]
\small
\setlength{\tabcolsep}{2pt} 
\renewcommand{\arraystretch}{1.1} 
\begin{tabular}{@{}ll@{}}
\hline
\textbf{Search Query}              	 & \textbf{Example of Title}                                                              \\
\hline
italy+travel+coronavirus 	 & Italy will PAY half the price of your hotel\_Travel Tourism News              \\
japan+travel+price       	 & Japan Foreign Arrivals Down 99.9\% In April \& To Cover 50 ...                \\
phones+missing            	 & A Thread from @ApostleKom: \textbackslash{}lq21 Million Chinese Cellphone ... \\
event 201+gates           	 & QAnon Supporters, Anti-Vaxxers Spread A Hoax Bill Gates ...                   \\
coronavirus+size+diameter & Carona Virus Updates ! Unicef Corona... - Jodhbir Singh ...                   \\
people+collapsing+china   & China Corona Virus Horror: Hospital Corridor of the Dead and Dying ...        \\
5g+coronavirus            	& New Study Suggests 5G Could Create Coronavirus Type ... - lbry.tv            \\
\hline
\end{tabular}
\caption{\label{tab:t1} \small Search queries submitted to Web Search Engine and meta information results. The results of a search query are often returned as a list of meta data and they may consist of web pages, images, and other types of files with helpful information.} 
\end{table}

The final dataset consisted of a collection of approximately 3350 news results (fake/misleading and trusted/\lq real'), gathered from 2084 different urls. All the news were published within a 7-month time interval, between January $20^{th}$ and July $28^{th}$, 2020. We chose this time interval as it covered the first CoViD-19 pandemic lockdowns proclaimed in Italy and in other countries \cite{FT}. 
Queries were selected based on topics (e.g., generic information on the new virus; pseudo-scientific therapies; conspiracy theories; travels; etc.) \citep{Naeem} that we were monitoring both on the Web (online newspapers) and Social Media, during the first lockdown period. We also looked at fact-checking websites, such as \href{https://www.politifact.com}{politifact.com} or \href{https://www.poynter.org}{poynter.com}, to check news and information credibility \citep{poynter}.

In order to reduce potential bias due to Search Engine optimisation, we had carefully planned our data collection as follows: 
\begin{itemize}
    \item we used a VPN to be more consistent with the WSE domain inspected and its results;
    \item in order to browse the Internet and query the WSE, we used a Private/Incognito window. This allowed us to prevent our browsing history from being stored and from biasing our results. By using Incognito/Private mode, we did not build a detailed picture of our online activity: in this way, all cookies were removed at the end of each session, i.e., we did not save any information about the pages we were visiting, avoiding to create customised results based on our search history. This process was repeated for each query and for each day of collection;
    \item we collected all the results, and not some of them (e.g., the first 2 pages or the top 10 results).
\end{itemize}

\begin{figure}[!htbp]%
    \includegraphics[width=15cm]{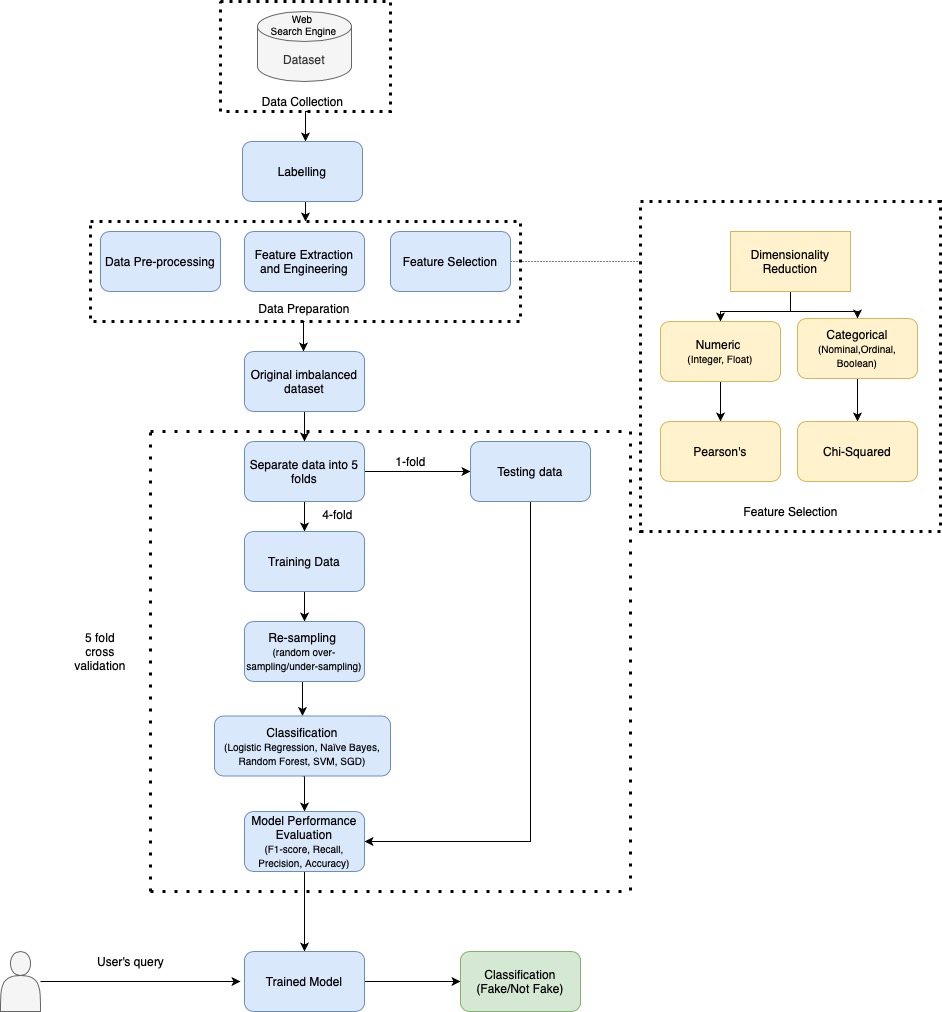}
    \caption{\small Machine Learning workflow for training algorithms and classification of results from Web Search Engines (WSE). A KNN imputation algorithm and MinMaxScalar were used to rescale variables. However, the percentage of missing values in the dataset was very low ($<1\%$). Since high correlation among features leads to redundancy of features and instability of the model, statistical tests, like Chi-Squared and Pearson's correlation coefficient, were used for feature selection. The corpus collected from WSEs was, therefore, pre-processed before being used as an input for training the models. Due to the class imbalance in the dataset, re-sampling techniques were applied to the training set only, during cross-validation (k-fold=5). Different classification models were then evaluated by scoring the classification outcomes from a testing set, in terms of the following performance metrics: $F_1$-score, Recall, Accuracy and Precision.} 
    \label{fig:2x}%
\end{figure}

Even if we dramatically reduced bias during our data collection, the results from WSE might be automatically biased by the WSEs we were querying because of their ranking systems, which sort results by relevance. We did not have any control on that but we tried to address this potential bias comparing results from different WSEs, also across different days.\\
Once we collected data, the labelling procedure was done manually and it consisted of assigning a binary class label indicating whether the news was real ($0$) or fake/misleading ($1$). In the binary fake news detection problem, fake news is usually associated with the positive class, since these are the news detected by the classifier. Data labelling process for training ML algorithms is not only critical but also time consuming. Because of the limited resources, we considered a limited sample size in our study, but big enough to be considered reliable and sufficiently large for binary detection \citep{Beleites}.\\
The ML workflow proposed in this study was implemented in Python 3.8. Its schematic representation is illustrated in \textbf{Figure} \ref{fig:2x}. 

\subsection{Data pre-processing and Feature Engineering}
\label{2b}
\begin{figure}[!htbp]%
    \centering
    \begin{tabular}{c}
        \includegraphics[width=7cm]{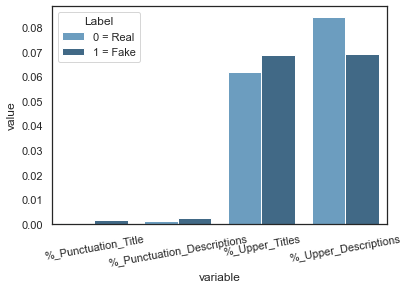}
        \small(A)
    \end{tabular}
    \begin{tabular}{c}
        \includegraphics[width=7cm]{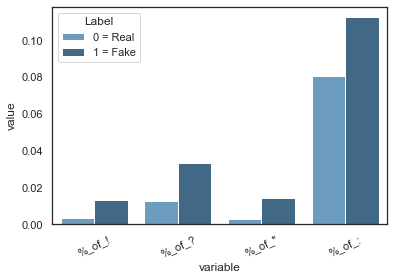}
        \small(B)
    \end{tabular}
    \caption{\small Frequency of: (A) punctuation characters and upper case words in meta titles and descriptions; (B) some specific punctuation characters (exclamation mark, question mark, quotes, colons). With the exception of upper case words that were used more frequently in meta descriptions, fake news has shown a higher percentage of punctuation in both title and description, probably by \lq over-dramatising' events; also, the use of upper case characters in titles appears more evident in misleading content.} %
    \label{fig:2}%
\end{figure}

In order to observe the most meaningful context words and to improve the performance of the classifiers, in the data pre-processing stage we removed all parts that were irrelevant, redundant and not related to the content: punctuation (with a few exceptions of symbols, like exclamation mark, question mark, and quotation marks) and extra delimiters; symbols; dashes from both titles and descriptions; stopwords \citep{Sarica}.\\ By following guidance and advice given by fact-checking websites (e.g., \href{https://www.factcheck.org/2016/11/how-to-spot-fake-news/}{factcheck.org}) and reputable outlet sources (e.g., \href{https://www.bbc.com/news/av/stories-51974040}{bbc.com}) on how to spot fake news, we looked at the presence of words in capital letters and at the excessive use of punctuation marks in both titles and descriptions. \textbf{Figure} \ref{fig:2} shows the frequency of specific punctuation characters (`!', `?',  ` `` ', `:') and upper case words in titles and descriptions for news labelled as fake (1) and real (0). 
It is notable that fake news differs much more for real one by the excessive use of punctuation, quotes, interrogatives, words in all capital letters, and exclamation mark to alert and urge people to read the news.

\begin{figure}[!htbp]
\centering
    \begin{tabular}{c}
        \includegraphics[width=7.5cm]{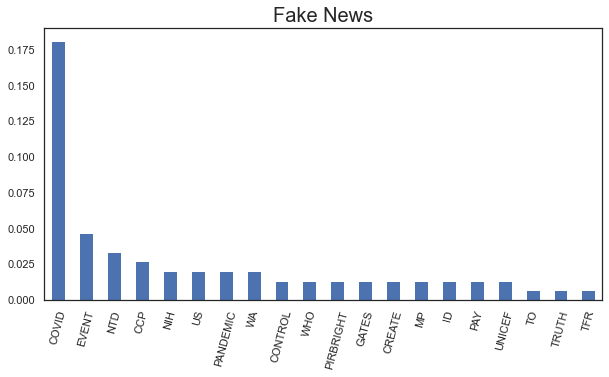}
        \small(A)
    \end{tabular}
    \begin{tabular}{c}
        \includegraphics[width=7cm]{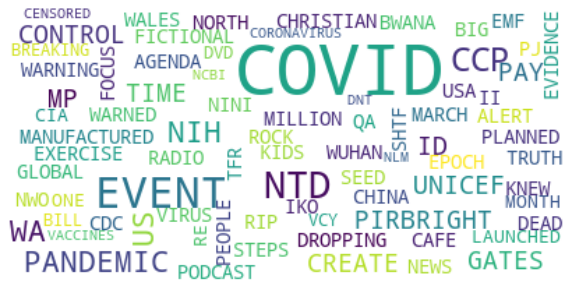}
        \small(B)
    \end{tabular}
    \begin{tabular}{c}
        \includegraphics[width=7.5cm]{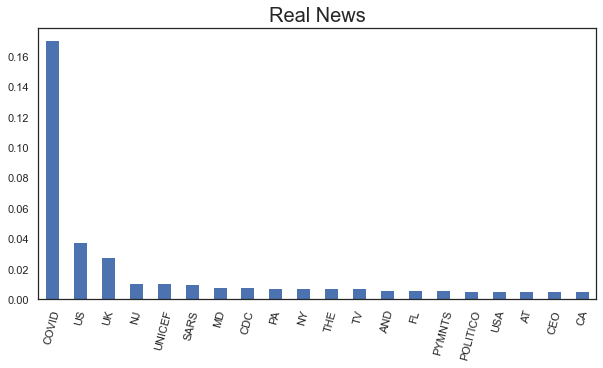}
        \small(C)
    \end{tabular}
    \begin{tabular}{c}
       \includegraphics[width=7cm]{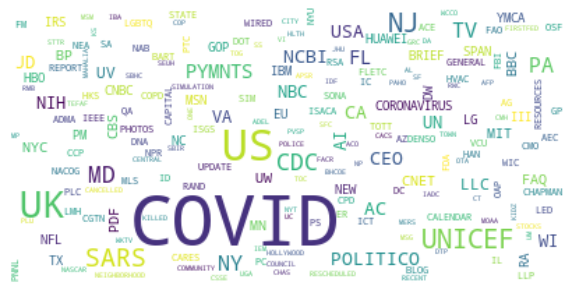}
       \small(D)
    \end{tabular}
    \caption{\small The 20 mostly used uppercase words in fake news (A) and in real news (C) datasets. In the sub-figures (B) and (D), respectively, the word clouds show words sized according to their weights in the datasets. The use of uppercase words is different between the two datasets: in real news, the use of uppercase words is more frequent to indicate acronyms, brands, organisations, while in fake news uppercase words emphasise feelings, creating alerts and potential warnings.}
    \label{fig:3}
\end{figure}

The frequency distributions in \textbf{Figure} \ref{fig:3} illustrate the top 20 uppercase words in the fake news and in the real news datasets. From the two histograms, we can derive an important information regarding the use of various uppercase words in the two news sets. It can be noticed, in fact, that in the real news dataset all uppercase words are more related to abbreviations (e.g., US, UK), acronyms (e.g., UNICEF), or organisations' name (NBC, NCBI), while in fake news dataset the use of uppercase letters highlights potential warnings (e.g., CONTROL, CREATE), capitalising on coronavirus fears and conspiracy theories. This highlights the different use of capitalising all characters in a word, an unusual habit for reporters working for trustworthy websites, who generally follow style-guidelines and journalistic convention.

In the feature engineering stage, which typically includes feature creation, transformation, extraction and selection), we used pre-training algorithms, such as Bag-of-Words (BoW) \citep{Zhang} and Term Frequency–Inverse Document Frequency (TF-IDF) \citep{Al, Ahmed}, for mapping cleaned texts (titles and descriptions) into numeric representations. Further features (length, counting, binary) were also extracted from URLs \citep{Zhu}.

Information on the age of domain names was gathered from both \href{http://web.archive.org}{Wayback Machine} and \href{https://who.is}{WHOIS} \citep{Li}, two tools that are crucial resources in the fight against fake news, as they allow users to see how a website has changed and evolved through time, gathering information on when the website was founded, on its country code top-level domain, and contact information.

\subsection{Re-sampling}
\label{2c}
Although various ML methods assume that the target classes have same or similar distribution, in real conditions this does not happen as data is unbalanced \citep{Luque}, with nearly most of the instances labelled with one class, while a few instances are labelled as the other one. Since we worked with real-world data \citep{Xie, Agrawal}, our dataset presented a high class imbalance with significantly less samples of fake news than real one. 
To address poor performance in case of unbalanced dataset, we used: 

\begin{itemize}
	\item \textit{minority class random over-sampling} technique, which consists in over-sizing the minority class by adding observations;
	\item \textit{majority class random under-sampling} technique, which consists in down-sizing the majority class by randomly removing observations from the training dataset. 
\end{itemize}

The re-sampling algorithms chosen depend on the nature of the data, specifically on the ratio between the two classes, fake/real. Although we had a class imbalance skewed (90:10), we could not treat our case as a problem of anomaly (or outlier) detection. In fact, in order to be considered such a case, we would have had a very skewed distribution (100:1) between the normal (real) and rare (fake) classes.\\
Although the choice of the number of folds is still an open problem, generally researchers choose a number of folds equal to 3 (less common), 5 or 10. We used a 5-fold cross validation due to the small size of our dataset, but enough to contain sufficient variation \citep{Lever}. Each fold was used once as a validation, while the k - 1 remaining folds formed the training set. This process repeatedly ran until each fold of the 5 folds were used as the testing set.
\\


\section{Results}
\label{3}
In this section we discuss features from URLs, the metrics used for evaluating models' performance and we report the classification results. 

\subsection{URLs analysis}
\label{3a}
\begin{figure}[!htbp]%
    \centering
  \includegraphics[width=\textwidth]{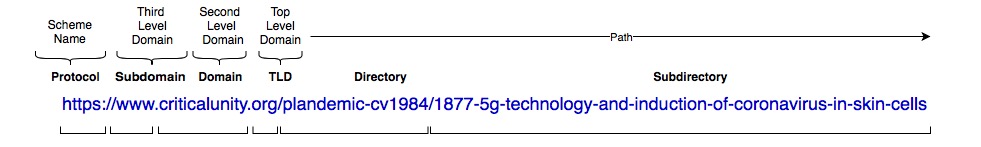} %
    \caption{\small URL's structure. A URL may contain mandatory (scheme, third-level, second-level, top-level domain) and optional (path) parts.} %
    \label{fig:5}%
\end{figure}

We analysed lexical and host-based features from 2084 distinct URLs. To implement lexical features, we used a Bag-of-Words of tokens in URL, where \lq /' , \lq ?', \lq .', \lq =', \lq \_', and \lq -'  are delimiters. We distinguished tokens that appear in the host name, path, top-level domain, using also the lengths of the host name and the URL as features \citep{Cho, Wejinya}. 
In \textbf{Table} \ref{tab:t2} we show all the features extracted from URLs. Word-based features were introduced as well, as URLs were found to contain several suggestive word tokens. An example of URL structure is shown indeed in \textbf{Figure} \ref{fig:5}, where it is possible to distinguish the following parts: 

\begin{itemize} 
	\item \textit{scheme}: it refers to the protocol, i.e., a set method for exchanging or transferring data, that the browser should use to retrieve any resource on the web. \textit{https} is the most secured version; 
	\item \textit{third-level domain}: it is the next highest level following the second-level domain in the domain name hierarchy. The most commonly used third-domain is \textit{www};
	\item \textit{second-level domain}: it is the level directly before the top-level domain. It  is generally the part of a URL that identifies the website's domain name;
	\item \textit{top-level domain}: it is the domain's extension. The most used TLD is \textit{.com}. The TLD can also give about geographic of a website, since each country has a unique domain suffix (e.g., .co.uk for UK websites). 
\end{itemize}


We used the Chi-Squared (${\chi}^2$) statistical test to assess the alternate hypothesis that the association we observed in the data between the independent variables (URL-feature) and the dependent variable (fake/not fake) was significant; specifically:

\begin{itemize}
    \item \textit{null hypothesis} ($H_0$): there is no significant association between the variables and the dependent variable (fake/not fake);
    \item \textit{alternate hypothesis} ($H_1$): there is an association between the variables and the dependent variable (fake/not fake). 
\end{itemize}

We set a significance level of 0.05 \citep{Di_Leo}: 

\begin{itemize}
    \item if the p-value was less than the significance level, then we rejected the null hypothesis and concluded that there was a statistically significant association between the variables; 
    \item if the p-value was greater than or equal to the significance level, we failed to reject the null hypothesis because there was not enough evidence to conclude that the variables were associated.  
\end{itemize}

The correlation-based feature selection (CFS) algorithm was used for evaluating the worth or merit of a subset of features, taking into account the usefulness of individual features for predicting the class label. \\
In order to check high correlations among independent variables, we also performed a multicollinearity test. Multicollinearity is indeed a common problem when estimating models such as logistic regression. In general, to simulate predictor variables with different degree of collinearity, the Pearson pairwise correlation coefficients were varied: an absolute correlation of greater than or equal to $0.7$ can be considered an appropriate indicator for strong correlation \citep{Vatcheva}. \\To measure the increase in the prediction error of the model, permutation importance feature was employed. The method is most suitable when the number of features is not huge as it is resource-intensive. This method can be also used for feature selection. In fact, it allows to select features based on their importance on the model. If there are features correlated, then the permutation importance will be low for all the correlated features.  The choice of permutation importance as extra method for feature selection was justified also by the use of different models, tree and not-tree based, respectively \citep{Gomez}.\\
These feature selection methods allowed us to select a small number of highly predictive features in order to avoid over-fitting.

\begin{table}[!htpb]
\footnotesize
\begin{tabular}{lllll}
\hline
\textbf{Feature} & \textbf{Feature Name}                                                                        & \textbf{Type}        		& \textbf{Fake News}	  & \textbf{Real News} \\
\hline
Domain &URL length                                                                             & continuous  		& $95.2\pm2.0 $        	  &  $82.1\pm0.7$    \\
&if domain starts with numbers                                                   		  & binary      		& 1.2\%              	  & 0.2\%        	     \\
&if domain is an IP address                                                       		  & binary     		& .             	  & .            	 \\
&if .com 												  & categorical 		& 85.0\%       	  & 73.3\%    	\\
&if .org												  & categorical 		& 8.4\%        	  & 15.0\%    	\\
&if .gov 												  & categorical 		& 0.3\%        	  & 4.6\%    	\\
&if .net                                                        		  				  & categorical 		& 2.3\%        	  & 2.3\%    	\\
&age                                                                               			 & continuous  		&  $2010\pm0.5$                & $2006\pm0.2$           \\
\hline
Word-based &if blogspot, blog, wordpress, blogger, ... is contained in the URL & binary  &  4.4\%                   & 1.7\%           	\\
&if facebook, twitter, instagram, ... is contained in the URL                   & binary      		& 8.7\%                    & 1.0\%            	\\
&if news press, journal, publisher, ... is contained in the URL               & binary      		& 10.2\%           	  & 7.1\%         	\\
&if coronavirus, virus, covid is contained in the URL                             & binary      		& 0.6\%               	  & 0.1\%          \\
\hline
Host-based & if http                                                                               & binary      		& 6.1\%           	  & 3.8\%         	\\
&if https                                                                          			 & binary     	 	& 63.7\%        	  & 57.5\%       \\
\hline
Lexical & dots count                                                                         	& continuous  		&  $1.73\pm0.04$       		 &  $2.07\pm0.02$  		\\
&semicolons count                                                               	    	& continuous  		& .              	 &  $0.002\pm0.002$\\
&ampersand count                                                               	     	& continuous  		& .              	 & 0.031\\
&slash count                                                                      		  	& continuous  		&  $5.19\pm0.08$         	 &  $5.195\pm0.029$        	\\
& hyphen count                                                                     	  	& continuous  		&  $7.8\pm0.3$                 &  $5.4\pm0.1$        \\
&underscore count                                                            		        & continuous  		& $0.3$                  & $0.2$       	      \\
&equal count                                                                       		 	& continuous 	 	&  $0.085\pm0.015$              &  $0.081\pm0.009$       	\\
&question mark count                                                       	         	& continuous  		&  $0.085\pm0.015     $         &  $0.053\pm0.004$       \\
&@ count                                                                          		  	& continuous  		&  $0.003\pm0.003$              &  $0.001\pm0.001$      	\\
& digit count                                 							& continuous 		&  $7.6\pm0.4$		 	 &  $5.7\pm0.1$		\\
\hline
\end{tabular}
\caption{\small List of domain, host-based, and lexical features, extracted from URLs. There are: 8 domain features; 4 word-based features, which show the presence of specific words in a URL; 2 host-based features; and the remaining 10 features are lexical-based and include special characters count or show the presence of digits in a URL.  The purpose of feature engineering was to find and explore new features to improve model performance for fake news detection. We used Chi-square statistics and a correlation-based feature selection (CFS) approach. If reported, the error is meant the standard error of the mean.}
\label{tab:t2}
\end{table}

\begin{figure}[!htbp]
  \centering
    \begin{tabular}{c}
        \includegraphics[width=0.43\textwidth]{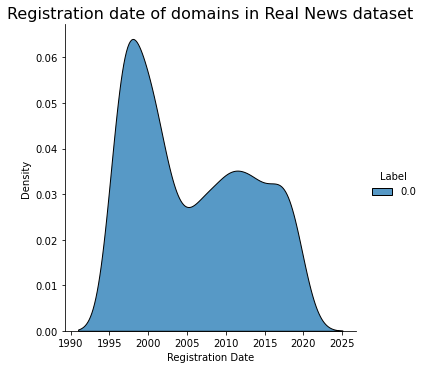}
        \small(A)
    \end{tabular} 
    \begin{tabular}{c}
        \includegraphics[width=0.43\textwidth]{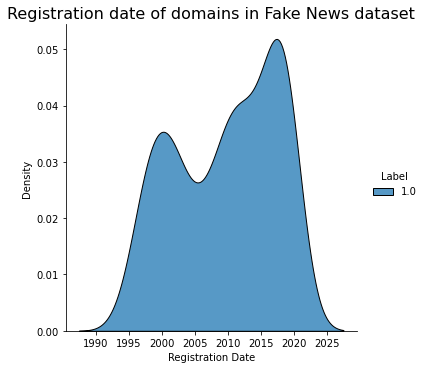}
        \small(B)
    \end{tabular}
  \caption
    {%
       \small Density distribution of domain name's age by class. The resulting distributions show a very clear negatively skew (fake news) versus a positively skew (real news).  Age of domain can be extracted from Wayback or WHOIS external services. Based on the analysis, we observed that fake news or misleading content is published or shared more likely by newer websites. %
      \label{fig:6}%
      }%
\end{figure}

In our fake news dataset, the average length of URL was 95.2 characters. This URL length is greater than the values found by \citep{Garera,Jeeva,Sankhwar} for phishing URLs. The number of hyphens (-) in the URL was found to be greater than 7 on average for fake news URLs. The most surprising result is in the number of dots in URL: fake news URLs in our dataset do not contain more than 2 dots on average.
In \textbf{Table} \ref{tab:t3} we can observe that  websites publishing fake news have generally newer domain name's age than websites publishing reliable news (\textbf{Table} \ref{tab:t3} and \textbf{Figure} \ref{fig:6}). 

{\renewcommand{\arraystretch}{1.5}
\begin{table}[!htbp]
\begin{tabular}{ cccccccc } 
\hline
& \textbf{Total} & \textbf{Mean} & \textbf{Median} & \textbf{10th percentile} & \textbf{90th percentile}\\
\hline
Websites publishing Fake news & 265 & 2010 & 2011 & 1998 & 2019 \\ 
Websites publishing Real news & 1865 & 2006 & 2004 & 1996 & 2017\\  
\hline
\end{tabular}
\caption{\label{tab:t3} \small Summary statistics of domain name's age (Wayback Machine) of distinct websites in our dataset.}
\end{table}
}
For validating the results shown in \textbf{Table} \ref{tab:t3}, we used Welch's t-test, which is usable independently of the data distribution thanks to the central limit theorem. The p-value ($4.195e^{-17}$) we got is less than the chosen significance level (0.05), therefore we reject the null hypothesis in support of the alternative.\\

\subsection{Performance metrics}
\label{3b}
In terms of model performance measurement, the decision made by the classifier can be represented as a $2 \times 2$ confusion matrix having the following four categories: 
\begin{itemize}
	\item True Positives (TP), i.e., number of positive instances that are correctly classified;
	\item False Positives (FP), i.e., number of misclassified positive instances;
	\item True Negatives (TN), i.e., number of negative instances that are correctly classified;
	\item False Negatives (FN), i.e., number of misclassified negative instances.
\end{itemize}

To evaluate the effectiveness of models, we used the metrics shown in \textbf{Table} \ref{tab:metrics} \citep{Bekkar}: 

{\renewcommand{\arraystretch}{2.0}
\begin{table}[!htbp]
\begin{tabular}{ ccc} 
\hline
\textbf{Metric} & \textbf{Formula} & \textbf{Description} \\
\hline
Precision & $\frac{TP}{TF+FP}$ & $\frac{\textrm{number of correctly classified positive instances}}{\textrm{total number of classified instances as positive}}$ \\
Accuracy & $\frac{TP+TN}{TP+FP+TN+FN}$ & $\frac{\textrm{number of correctly classified instances}}{\textrm{total number of classified instances}}$\\
Recall & $\frac{TP}{TP+FN}$ & $\frac{\textrm{number of correctly classified positive instances}}{\textrm{actual number of positive instances}}$ \\
$F_1$-Score & $2\times \frac{Precision\times Recall}{Precision+Recall}$ & $\textrm{harmonic mean of precision and recall}$ \\
\hline
\end{tabular}
\caption{\label{tab:metrics} \small Evaluation metrics based on confusion matrix. Both $F_1$ score and Recall are good metrics for the evaluation of imbalanced data.}
\end{table}

Both $F_1$ score and Recall are good metrics for the evaluation of imbalanced data.

\subsection{Model Evaluation}
\label{3c}
Since we are dealing with imbalanced data, the predictive, accuracy represents a misleading indicator, as it reflects the underlying class distributions, making it difficult for a classifier to perform well on the minority class \citep{Al}. For this reason, we used $F_1$ score \citep{Luque, Jeni} and Recall metrics, as the higher the value assumed by these metrics, the better the class of interest is classified.\\
\textbf{Table} \ref{tab:t4} shows the evaluation metrics for all the classifiers we considered. It can be noticed that the classification metrics depend on the type of classifier and on the extracted features used for the classification. 
Logistic Regression with BoW model was the most effective classifier when we oversampled the data, reaching the highest $F_1$-score (71\%), followed by Na\"ive Bayes with BoW model (70\%), and SVM with TF-IDF (69\%). \\When we used the under-sampling technique and removed instances from the majority class, the score of the classifier models was very poor compared to over-sampling technique. SGD with TF-IDF and Na\"ive Bayes with TF-IDF and BoW came out the worst with $F_1$ scores of 34\%, 35\%, and 37\%, respectively. From \textbf{Table} \ref{tab:t4}, only Random Forest classifier got a $F_1$-score greater than 50\%, unlike the other classifiers when the under-sampling algorithm was applied, though the Precision metric results very poor.

{\renewcommand{\arraystretch}{1.2}
\begin{table}[!htbp]
\begin{tabular}{ ccccccc } 
\hline
\textbf{Re-Sampling} & \textbf{Classifier} & \textbf{Pre-Processing} & \textbf{Precision} & \textbf{Accuracy} & \textbf{Recall} & \textbf{$F_1$-Score}\\
\hline
& \textbf{Na\"ive Bayes} & \textbf{BoW}  & \textbf{0.64} & \textbf{0.92} & \textbf{0.78} & \textbf{0.70}\\ 
& \textbf{Logistic Regression} & \textbf{BoW} & \textbf{0.73} & \textbf{0.93} & \textbf{0.68} & \textbf{0.71} \\ 
& SVM & BoW & 0.73 & 0.93 & 0.63 & 0.68 \\ 
& SGD & BoW  & 0.70 & 0.92 & 0.62 & 0.66 \\ 
Over & Random Forest & BoW  & 0.82 & 0.92 & 0.42 & 0.56 \\ 
& Na\"ive Bayes & TF-IDF  & 0.38 & 0.82 & 0.88 & 0.53 \\ 
& Logistic Regression & TF-IDF  & 0.68 & 0.92 & 0.67 & 0.67 \\ 
& \textbf{SVM} & \textbf{TF-IDF}  & \textbf{0.79} & \textbf{0.94} & \textbf{0.60} & \textbf{0.69} \\ 
& SGD & TF-IDF   & 0.60 & 0.91 & 0.64 & 0.62 \\ 
& Random Forest & TF-IDF   & 0.76 & 0.93 & 0.58 & 0.66 \\ 
\hline
& Na\"ive Bayes & BoW  & 0.23 & 0.64 & 0.91 & 0.37 \\  
& Logistic Regression & BoW  & 0.38 & 0.83 & 0.79 & 0.51 \\ 
& SVM & BoW & 0.34 & 0.80 & 0.79 & 0.47 \\ 
& SGD & BoW  & 0.26 & 0.71 & 0.83 & 0.39 \\ 
Under & \textbf{Random Forest} & \textbf{BoW}  & \textbf{0.46} & \textbf{0.87} & \textbf{0.74} & \textbf{0.57} \\ 
& Na\"ive Bayes & TF-IDF & 0.22 & 0.62 & 0.89 & 0.35 \\ 
& Logistic Regression & TF-IDF & 0.34 & 0.79 & 0.84 & 0.48 \\ 
& SVM & TF-IDF & 0.30 & 0.75 & 0.85 & 0.44 \\ 
& SGD & TF-IDF & 0.22 & 0.64 & 0.83 & 0.34 \\ 
& \textbf{Random Forest} & \textbf{TF-IDF} & \textbf{0.48} & \textbf{0.88} & \textbf{0.68} & \textbf{0.57} \\  
\hline
\end{tabular}
\caption{\label{tab:t4} \small Evaluation metrics (Precision, Accuracy, Recall, and $F_1$ score) for all the classifiers, considering only textual features selection. Bold values represent the highest performance results based on $F_1$ score. We used different metrics for evaluating ML classifiers with BoW and with TF-IDF. As shown in this table, Logistic Regression classifier performed better with $F_1$  score $0.71$ compared to other classifiers using BoW in case of over-sampled data. Similarly, SVM classifier performed better using TF-IDF in case of over-sampled data. Due to the small size of the dataset, under-sampling technique was not well-suited for the classifiers, as the results showed that the classifiers performed very poorly.}
\end{table}

\textbf{Figure} \ref{fig:7} shows a comparison of the classifiers using different feature extraction techniques (BoW and TF-IDF) based on $F_1$-score metric (\textbf{Table} \ref{tab:t4}). 

\begin{figure}[!htbp]
  \centering
    \begin{tabular}{c}
        \includegraphics[width=0.35\textwidth]{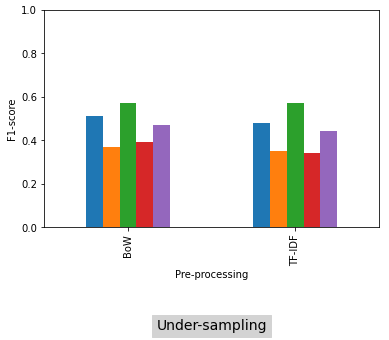}
        \small(A)
    \end{tabular}
    \begin{tabular}{c}
        \includegraphics[width=0.48\textwidth]{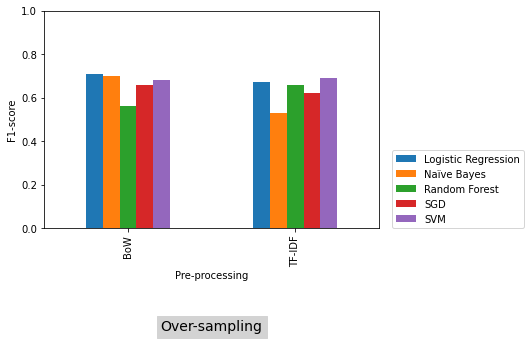}
        \small(B)
    \end{tabular}
  \caption
    {%
      \small Comparison of classification $F_1$ score metric for the models obtained by the two feature extraction methods (BoW and TF-IDF) for under-sampled (A) and over-sampled (B) data. $F_1$ score is used in the case of imbalanced datasets.}
      \label{fig:7}%
      
\end{figure}

Based on the analysis we performed on Section \ref{3a}, we observed a positive influence on the $F_1$-score and Recall metrics (\textbf{Figure} \ref{fig:8}) in some ML classifiers, after including the most relevant features extracted from URLs. 
As shown in \textbf{Table} \ref{tab:t5}, the implementation of new features extracted from URLs successfully assisted the classifiers, by improving their performance.  

\renewcommand{\arraystretch}{1.2}
\begin{table}[!htbp]
\begin{tabular}{ ccccccc } 
\hline
\textbf{Re-Sampling} & \textbf{Classifier} & \textbf{Pre-Processing} & \textbf{Precision} & \textbf{Accuracy} & \textbf{Recall} & \textbf{$F_1$-Score}\\
\hline
     & 	\textbf{Na\"ive Bayes}         & \textbf{BoW}           & \textbf{0.76}  & \textbf{0.96} & \textbf{0.86} & \textbf{0.81}   \\
     & Logistic Regression 		 & BoW           & 0.77  & 0.94 & 0.67 & 0.72  \\
     & SVM                			 & BoW           & 0.81  & 0.94 & 0.64 & 0.71   \\
     & SGD               			 & BoW           & 0.79  & 0.94 & 0.64 & 0.71   \\
Over     & Random Forest       	 & BoW           & 0.90  & 0.94 & 0.50 & 0.64   \\
     & Na\"ive Bayes        		 & TF-IDF        & 0.34  & 0.79 & 0.97 & 0.50   \\
     & Logistic Regression 		 & TF-IDF        & 0.72  & 0.94 & 0.78 & 0.75   \\
     & \textbf{SVM}                 	 & \textbf{TF-IDF}        & \textbf{0.80}  & \textbf{0.95} & \textbf{0.78} & \textbf{0.79}   \\
     & SGD                 			 & TF-IDF        & 0.82  & 0.95 & 0.71 & 0.76   \\
     & Random Forest       		 & TF-IDF        & 0.80  & 0.94 & 0.65 & 0.72\\  
\hline
\end{tabular}
\caption{\label{tab:t5} \small Evaluation metrics (Precision, Accuracy, Recall, and $F_1$ score) for all the classifiers after URL features selection. Bold values represent the highest performance results based on $F_1$-score. As shown in this table, Na\"ive Bayes classifier performed better with $F_1$  score $0.81$ compared to other classifiers using BoW in case of over-sampled data. Similarly, SVM classifier performed better using TF-IDF with $F_1$  score $0.79$. It is evident that adding the URL features improved the model's performance. }
\end{table}

A visual inspection of metrics by model, before and after adding URL features in our ML classifiers, is illustrated in \textbf{Figure} \ref{fig:8}. The results verify the effectiveness of introducing URL features, with values approximately above 0.70 for the two types of pre-processing. Before URL features selection, the highest $F_1$-score was 0.71.

\begin{figure}[!htbp]
  \centering
  \includegraphics[width=0.9\textwidth]{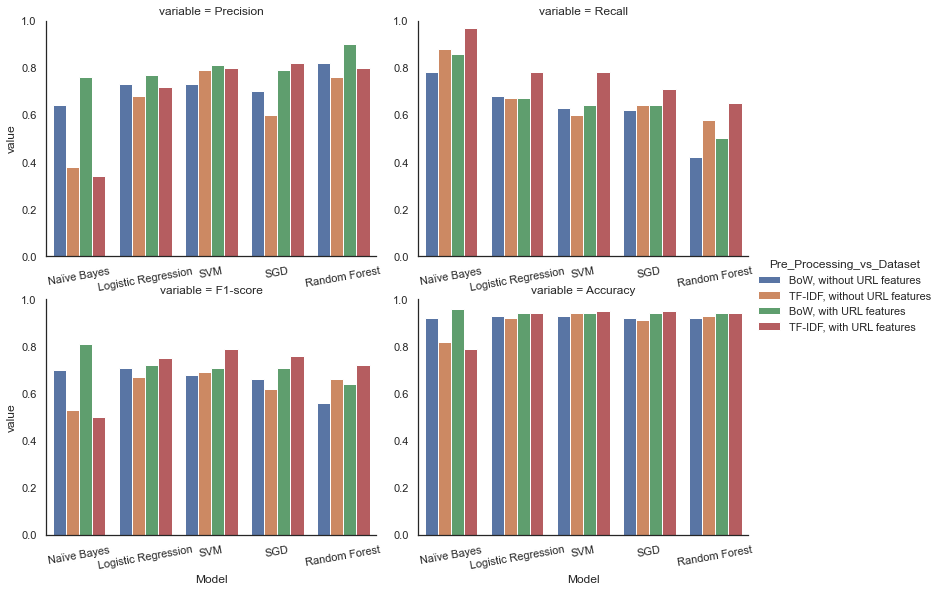}%
  \caption
    {%
      \small Comparison of metrics (precision, accuracy, recall, and $F_1$-score) by model, before and after adding URL features in our ML classifiers. Overall, the results verify the effectiveness of introducing URL features. This is more evident in the classifiers that use TF-IDF. 
      \label{fig:8}%
      }%
\end{figure}

}

\section{Discussion}
\label{4}
In binary classification problems, class imbalance represents an open challenge as real-word datasets are usually skewed. One issue involves the determination of the most suitable metrics for evaluating model performance. $F_1$ score, defined as the harmonic mean of Precision and Recall (Section \ref{3b}), has been commonly used to measure the level of imbalance. Our data had a significantly high level of imbalance (majority class, i.e., real news, was approximately 90\% of our dataset, and minority class, i.e., fake news, represented only 10\% of the dataset). A way to address and mitigate class imbalance problem was data re-sampling, which consists of either over-sampling or under-sampling the dataset. Over-sampling the dataset is based on rebalancing distributions by supplementing artificially generated instances of the minor class (i.e., fake news). On the other hand, under-sampling method is based on rebalancing distributions by removing instances of the majority class (i.e., real news). By under-sampling the majority class, we had to reduce the sample size, which resulted too small for training models, causing poor performance. By over-sampling data, we instead noticed better results in terms of both Recall and $F_1$ score metrics, boosting up the model performance. \\
We compared models based on popular feature representations, such as BoW and TF-IDF. After over-sampling data, the evaluation metrics returned results with $F_1$-score over 70\% for both Logistic Regression and Na\"ive Bayes classifiers with BoW. \\ 
In order to further improve the results, we decided to focus on news sources as well, exploring and selecting URL features that have displayed high impact in various studies \citep{Ranganayakulu, Sonowal, Li}. \\
In fact, just like phishing attacks (e.g., suspicious e-mails or malicious links), fake news continues to be a top concern, as they still spread across the Web and will continue to spread until everyone understands how to spot them.\\ A comparison between phishing websites and websites that deliberately have published fake news is shown in \textbf{Table} \ref{tab:t6}. It is evident that websites that publish and share misleading content have generally URLs with identifiable features (\textbf{Table} \ref{tab:t2}), like malicious URLs.

\renewcommand{\arraystretch}{1.2}
\begin{table}[]
\begin{tabular}{ccccc}
\hline
\textbf{Type}      & \textbf{Not-reliable website}    & \textbf{Reliable website}  & \textbf{Registration Date - NR} & \textbf{Registration Date - R}                                                                          																									       \\ \hline
phishing		 & paypal--accounts.com             & paypal.com                       & NA 						& 1999                              \\ 
phishing  		 & fr.facebok.com                         & facebook.com                   & NA 						& 1997                              \\ \hline
fake news 	 & cnn-trending.com                     & cnn.com           		     & 2017 					& 1993        			\\ 
fake news 	 & fox-news24.com 			   & foxnews.com        	     & 2018 					& 1995       			\\ 
fake news 	 & abcnews.com.co			   & abcnews.com        	     & 2016 					& 1995       			\\ 
fake news 	 & ilfattoquotidaino.it			   & ilfattoquotidiano.it             & 2016 					& 2009       			\\ \hline
\end{tabular}
\caption{\label{tab:t6} \small Comparison of not-reliable (NR) and reliable (R) websites.  Alongside lexical features, domain name's registration date (extracted from WHOIS) can be relevant to spot fake websites. }
\end{table} 

As shown in \textbf{Table} \ref{tab:t6}, phishing carries out also by \textit{typosquatting} domain, i.e., by registering a domain name that is extremely similar to that of an existing popular one. In the past few years, various online websites have been created to imitate trustworthy websites in order to publish misleading and fake content: for example, \textit{abcnews.com} (registered on 1995) and \textit{abcnews.com.co} (registered ahead of the 2016 US election); or \textit{ilfattoquotidiano.it} (registered on 2009) and \textit{ilfattoquotidaino.it} (registered on 2016). \\
One of the most relevant URL features was certainly registration date. In our dataset, the average age of domain name of websites publishing fake news was 2008, while that one of websites publishing real news was 2004 (\textbf{Table} \ref{tab:t3}). Most of websites publishing fake news are, therefore, newer than websites which spread reliable news. This was in line with our expectation, i.e., that websites publishing reliable news are typically older, having more time to build reputation, while those ones that publish fake news and misleading content are likely unknown websites created more recently. \\
The effects on the other features extracted from URLs had also a positive impact on the detection problem. By using correlation matrix heatmap and looking at findings from other research works, we selected features that most affected the target variable.
Like in phishing, websites or blogs that publish and share fake news may contain special symbols (such as @ and \&) to obfuscate links and trick readers into thinking that the URL leads to a legitimate website. For example, abcnews.com.co is a fake website, where the use of dots is for adding an extension (i.e., .co). On the other hand, the proportion of http and https did not provide relevant information, as https secured protocol now is commonly used. 
News by TLD showed that the most popular TLDs are .com (85\% in fake news dataset; 73.3\% in real news dataset) and .org. (8.4\% in fake news dataset; 15\% in real news dataset) (\textbf{Table} \ref{tab:t2}).  Furthermore, large numbers of digits and hyphens (greater than 7 on average) were found within URLs in the fake news dataset, making it possible to compare with results obtained from the analysis of malicious URLs  \citep{Wang, Palaniappan}.  
\\
By entering the selected URL features the model, Na\"ive Bayes classifier with BoW achieved $F_1$ score of 81\%, while SVM with TF-IDF got 79\%, significantly exceeded results based on features built from lexical representations of the text (titles and descriptions) only.

Based on the achieved result, we concluded that the use of URL features increased the performance of models.  \\
In terms of challenges, the class imbalance of real-world data and the limited accessibility of high-quality labelled dataset are two of the major ones. The use of ML classification models in fake news detection still appears more challenging in realistic situations, especially on Web Search Engines, where metadata information from thousands websites are collected.

\begin{table}[]
{\fontsize{8}{8}\selectfont 
\begin{tabular}{llll}
\hline
\textbf{Date}       & \textbf{URL}                                    & \textbf{Title}                                                                                                              & \textbf{Description}                                                                                                                                                             \\ \hline
2020-01-27 & \href{https://prepareforchange.net/2020/01/27/event-201-bill-gates-world-economic-forum-simulated-coronavirus-outbreak-6-weeks-before-first-case-in-wuhan/}{prepareforchange.net} & \thead{The “Event” is the moment of the\\ “Compression Breakthrough” on earth} & \thead{It is a cosmic EVENT HORIZON created \\by big solar waves reaching the\\ Earth from the Galactic Central causing \\the activation of “The Compression ...}                              \\ 
2020-03-04 & \href{http://www.drugtodayonline.com/medical-news/latest/10597-sun-exposure-washing-hands-kills-corona-virus-unicef.html}{www.drugtodayonline.com}                                   & \thead{Sun exposure, washing hands kills\\ corona virus: Unicef} & \thead{The Unicef has said that corona \\virus is large in size where the cell\\ diameter is 400-500 micro and for this \\reason any mask prevents its entry. The virus \\does not settle in the air but\\ is grounded, so it is not transmitted by air.} \\ 
2020-03-20 & \href{https://politicalfilm.wordpress.com/2020/03/20/event-201-october-18-2019/}{politicalfilm.wordpress.com}  & \thead{Event 201: October 18, 2019 |\\ Political Film Blog}                                                & \thead{John Hopkins / Bill Gates Foundation ran a \\CORONAVIRUS event in October. How\\ prescient???They claim it originated in pigs, \\contradicting the wild animal narrative \\that currently dominates. The bioweapons\\ question is nowhere to be seen. }        \\ 
2020-03-22 & \href{https://www.theepochtimes.com/the-closing-of-21-million-cell-phone-accounts-in-china-may-suggest-a-high-ccp-virus-death-toll\_3281291.html}{www.theepochtimes.com}           & \thead{21 Million Fewer Cellphone Users\\ in China May Suggest a High CCP\\ Virus Death Toll} & \thead{The number of Chinese cellphone users\\ dropped by 21 million in the past\\ three months, Beijing authorities\\ announced on March 19.}     \\ 
2020-03-28 & \href{https://beforeitsnews.com/conspiracy-theories/2020/03/coronavirus-follow-the-money-and-the-players-2516522.html}{beforeitsnews.com}  & \thead{Coronavirus: Follow The Money... \\and the Players!}  & \thead{Coronavirus:  Follow the Money … \\and the Players! – The problem \\with the Deep State Shadow Government \\– America, contracted out!}                          \\ 
2020-04-04 & \href{https://productivityhub.org/2020/04/04/bill-gates-calls-for-a-digital-certificate-to-identify-who-received-covid-19-vaccine/}{productivityhub.org}   & \thead{BILL GATES CALLS FOR A \\“DIGITAL CERTIFICATE” TO \\IDENTIFY WHO RECEIVED \\COVID-19 VACCINE} & \thead{In October 2019 (only a few \\months before the apparition of \\COVID-19) the Bill and Melinda Gates \\Foundation (in cooperation with \\the World Economic Forum) hosted \\Event 201, a 3.5-hour table-top simulation \\of a global pandemic.}         \\ \hline
\end{tabular}
}
\caption{\label{tab:t7} \small Example of fake news in our dataset. CoViD-19 pandemic has been resulted in misleading information, false claims and pseudo-scientific therapies, regarding the diagnosis, treatment, prevention, origin and spread of the virus. Conspiracy theories have been widespread too, mostly online through blogs, forums and social media.}
\end{table}


Furthermore, as in phishing attacks, who writes fake news and misleading content constantly looks for new and creative ways to fool users into believing their stories involve a trustworthy source. This makes necessary to keep models continuously updated as fake news is becoming more and more sophisticated and difficult to spot.  Also, misleading contents vary greatly and change over time: therefore, it is essential to investigate new features. 


\section{Conclusion}
\label{5}

In this study, we analysed meta data information extracted from Web Search Engines, after submitting specific search queries related to the CoViD-19 outbreak, simulating a normal user's activity.
By using both textual and URLs properties of data, we trained different Machine Learning algorithms with pre-processing methods, such as Bag-of-Words and TF-IDF. In order to deal with class imbalance due to real-world data, we applied re-sampling techniques, i.e., over-sampling of fake news and under-sampling of real news. While over-sampling technique allowed us to get satisfactory results, the under-sampling method was not able to increase model performance, showing very poor results due to the small sample size. \\ 
Although news has some specific textual properties which can be used for its classification as fake or real, when we look at search results (titles, snippets, and links), some additional pre-processing can be used to obtain some specific extra features for fake news detection on WSEs. While text features are related to news content, gathered from both titles and snippets, URL features are based on the source websites returned as search results on WSEs.\\ While most previous studies focused on fake news detection in Social Media, relying on data which can be directly gathered from the text (e.g., tweets) and from the usage of URLs for improving source credibility, our proposed approach goes further and analyse URL-features of the source of information itself. \\We believe indeed that URL pattern analysis via phishing detection techniques can enhance ML algorithms ability to detect and mitigate the spread of fake news across the World Wide Web. Checking the source is, indeed, one of the most common advice that fact-checking websites give to online readers \citep{fact}.\\
The results from this study suggest that information on URLs, extracted by using phishing techniques (e.g.,  number of digits, dots and length of the URL), could provide indications to researchers regarding a number of potentially useful features that future fake news detection algorithms might have or develop in order to bring out further valuable information on websites containing mostly false content and improve the model performance. \\
The analysis of fake news which spreads on the Web might have, however, a potential limitation, due to Search Engine optimisation. In this study we proposed a possible solution to address it. In fact, although Search Engine results might be customised based on online user location and user's search history, in order to reduce bias due to prior searching on the WSEs, it would be helpful to change settings preferences, delete cache, cookies, search history or use Incognito/Private windows. Furthermore, the use of proxies (or VPN) could allow to search queries on WSEs being location independent.\\
In terms of future research on fake news detection, we believe that techniques commonly used for malicious URLs detection should also be considered for fake news detection: this would mean building classifiers based not only on traditional lexical and semantic features of texts, but also on lexical and host-based features of the URL.
\\As future work, we therefore plan to construct more discriminative features to detect fake content, by profiling malicious sources of information based on domains, investigating in more detail, with additional performance metrics such as Net Reclassification Index (NRI), the improvement in prediction performance gained by adding a marker to the set of baseline predictors, in order to facilitate designing even better classification models for fake news detection.



\clearpage

\clearpage
\bibliographystyle{unsrt}
\bibliography{detection_of_fake_news_on_covid_19_on_WSE} 

\end{document}